\documentclass[letterpaper, 10 pt, journal, twoside]{IEEEtran}  

\IEEEoverridecommandlockouts                              

\usepackage{cite}
\usepackage{graphics} 
\usepackage[dvipdfmx]{graphicx}
\usepackage{epsfig} 
\usepackage{mathptmx} 
\usepackage{times} 
\usepackage{amsmath} 
\usepackage{amssymb}  
\usepackage{multirow}
\usepackage{comment}
\usepackage{color}
\usepackage{bm}
\usepackage{subfigure}
\usepackage{threeparttable}
\usepackage{tabularx}
\usepackage{booktabs}
\usepackage{float}
\usepackage{url}
\usepackage{bmpsize}
\usepackage{siunitx}

\title{
Bidirectional Trajectory Computation \\ for Odometer-Aided Visual-Inertial SLAM
}

\author{Jinxu Liu, Wei Gao* and Zhanyi Hu
\thanks{Manuscript received: September 24, 2020; Revised December 22, 2020; Accepted January 21, 2021.
}
\thanks{This paper was recommended for publication by Editor Sven Behnke upon evaluation of the Associate Editor and Reviewers' comments. This work was supported in part by the National Key R\&D Program of China (2016YFB0502002), and in part by the Natural Science Foundation of China (61991423, 61872361). (\emph{*Corresponding Author: Wei
Gao}) .}
\thanks{
All authors are with National Laboratory of Pattern Recognition, Institute of Automation, Chinese Academy of Sciences, and with School of Artificial Intelligence, University of Chinese Academy of Sciences, China. {\tt\small \{jinxu.liu, wgao, huzy\}@nlpr.ia.ac.cn}
}}

\begin{document}

\IEEEoverridecommandlockouts
\IEEEpubid{\makebox[\columnwidth]{978-1-5386-5541-2/18/\$31.00~\copyright2021 IEEE \hfill} \hspace{\columnsep}\makebox[\columnwidth]{ }}
\maketitle
\IEEEpubidadjcol
\markboth{IEEE Robotics and Automation Letters. Preprint Version. Accepted January, 2021}
{Liu \MakeLowercase{\textit{et al.}}: Bidirectional Trajectory Computation for Odometer-Aided Visual-Inertial SLAM}

\begin{abstract}
Odometer-aided visual-inertial SLAM systems typically have a good performance for navigation of wheeled platforms, while they usually suffer from degenerate cases before the first turning. In this paper, firstly we perform an observability analysis w.r.t. the extrinsic parameters before the first turning, which is a complement of the existing results of observability analyses. Secondly, inspired by the above observability analyses, we propose a bidirectional trajectory computation method, by which the poses before the first turning are refined in the backward computation thread, and the real-time trajectory is adjusted accordingly. Experimental results prove that our proposed method not only solves the problem of the unobservability of accelerometer bias and extrinsic parameters before the first turning, but also results in more accurate trajectories in comparison with the state-of-the-art approaches.
\end{abstract}

\section{INTRODUCTION}
%
%
%
%
\IEEEPARstart{V}{isual-inertial} SLAM (VI-SLAM) and visual-inertial odometry (VIO) approaches have received great attention from the researchers in recent years. They can be applied on mobile devices, micro aerial vehicles (MAVs), ground robots and passenger cars for localization and perception. Filtering-based methods such as MSCKF \cite{mourikis2007multi} and optimization-based methods such as OKVIS \cite{leutenegger2015keyframe} make up the two categories of state estimation methods. In general, filter-based approaches have better efficiency while optimization-based methods enjoy higher accuracy \cite{Huang2019Review}. The optimization-based approaches \cite{leutenegger2015keyframe},\cite{qin2017vins} typically optimize a limited number of current states to limit the amount of computation, and use marginalization to make use of previous information to better estimate the current states.

For wheeled platforms such as robots and passenger cars, the accuracy of visual-inertial navigation can be dramatically improved with the aid of wheel encoders \cite{wu2017vins},  \cite{liu2019visual}, \cite{Zhang2019vision}, \cite{zhang2019localization}, \cite{kang2019vins}, \cite{dang2018tightly}, \cite{ma2019ack}, \cite{quan2019tightly}. In some methods such as \cite{wu2017vins}, \cite{li2017gyro}, \cite{Zhang2019vision},\cite{zhang2019localization},\cite{kang2019vins}, \cite{dang2018tightly}, \cite{ma2019ack} and \cite{zuo2019visual}, the IMU measurements and wheel encoder readings are pre-integrated individually. This type of approaches either need at least two wheel encoders or require the front wheel angle measurement for pre-integration, and have to deal with the problem that the angular velocity provided by the sensors on wheels is always within the ground plane. Other methods such as \cite{quan2019tightly}, \cite{Ye2019robust} and \cite{liu2019visual} jointly pre-integrates the angular velocity from IMU and the linear velocity from wheel encoder. This type of approaches can still work when only one wheel encoder is available, and the uneven terrain does not have an impact on the performance of these approaches theoretically.
In this paper we use \emph{wheel encoder} and \emph{odometer} to denote the same thing.

However, for applications on ground vehicles, VI-SLAM and VIO approaches often suffer from degenerate cases, even with the aid of wheel encoders. \cite{wu2017vins} points out two degenerate cases under special motions. Firstly, the scale is unobservable when the platform moves with constant local linear acceleration. Secondly, the roll and pitch angles are unobservable when the platform has no rotational motion. Both the cases are related to the accelerometer bias which can not be correctly estimated under the above special motions. \cite{wu2017vins} also proves that the first degenerated case can be eliminated with the use of wheel encoder. However, it can be drawn that the second degenerate case still exists in such a case through derivation, which is straightforward and will be presented briefly in our technical report\footnote{\url{https://arxiv.org/abs/2002.00195v3}}. Besides, the experimental results in \cite{liu2019visual} also indicate that the accelerometer bias can not be correctly estimated until the first turning with the use of wheel encoder. In addition to the accelerometer bias, some of the extrinsic parameters can not be correctly estimated as well, when the platform has no rotational motion. \cite{yang2019degenerate} have proved that the translational component of camera-IMU extrinsic parameters is unobservable in a VIO system when the platform undergoes pure translation, and \cite{lee2020visual} have proved that the translational component of IMU-odometer extrinsic parameters is unobservable in an odometer-aided VIO system when the platform undergoes pure translation. However, our proposed approach adopts a sensor fusion scheme that is different from \cite{lee2020visual}, i.e. in our proposed approach IMU and wheel encoder measurements are fused in the pre-integration stage. One wheel encoder is sufficient to aid the VI-SLAM in our proposed approach, while at least two wheel encoders are required in \cite{lee2020visual}. Hence the observability analysis on our proposed approach is necessary. Furthermore, both \cite{yang2019degenerate} and \cite{lee2020visual} focus on the unobservable directions under pure translation, but neither of them have analyzed the observability under pure translation along a straight line, which is a common case and renders another direction in extrinsic parameters unobservable. In Section \ref{sec:observability} of this paper, we will give an observability analysis for the seven unobservable directions caused by the special motion pattern in extrinsic parameters for an odometer-aided VI-SLAM system, under the circumstance that the platform undergoes pure translation along a straight line. Thanks to the employment of marginalization, the optimization-based VI-SLAM approaches can make use of previous information collected since beginning. Therefore, once the platform performs rotational motion such as making a turn, the accelerometer bias and extrinsic parameters will be correctly estimated from then on. Nevertheless, before the first turning, the inaccuracy which results from the incorrectly estimated accelerometer bias and extrinsic parameters remains a problem for odometer-aided VI-SLAM approaches.

To relieve the inaccuracy before the first turning, \cite{liu2019visual} proposes to keep the extrinsic parameters constant during nonlinear optimization until the platform has made a turn and the estimation of accelerometer bias has reached convergence. Furthermore, we may further keep the accelerometer bias constant as zero before the first turning to relieve the inaccuracy caused by the incorrectly estimated accelerometer bias. Some initialization approaches for VI-SLAM also deal with unobservability when initial motion does not render all parameters observable. \cite{campos2019fast} forces accelerometer bias to be close to zero, and \cite{campos2020inertial} adds a prior with the mean value of zero w.r.t. the  accelerometer bias. Extrinsic parameters are not calibrated during VI-SLAM initialization. However, since the accelerometer bias is actually not zero and the extrinsic parameters may not be very accurate, the accuracy of trajectory before the first turning may still be deteriorated, especially for outdoor scenes where the vehicle is likely to travel a long distance before the first turning. 

By contrast, in this paper we propose a bidirectional trajectory computation approach to make the estimated poses before the first turning as accurate as those after the first turning. In short, after the first turning, we additionally create a backward computation thread to recalculate the poses from the first turning back to the starting point. In this way, the accuracy of estimated poses before the first turning do not suffer from the lack of rotation anymore, because the backward computation makes use of the information obtained in the first turning. Also note that by means of bidirectional trajectory computation, we can obtain more accurate overall trajectory in real time, because every time one of the poses before the first turning is updated by the backward computation thread, the real-time trajectory is also adjusted accordingly. In the following, we provide the observability analysis for the extrinsic parameters in Section  \ref{sec:observability} and describe the proposed bidirectional trajectory computation method in Section \ref{sec:method}.

\section{PRELIMINARIES ON ODOMETER-AIDED VISUAL-INERTIAL SLAM}

The proposed bidirectional trajectory computation method is based on the odometer-aided VI-SLAM approach \cite{liu2019visual}, which is a tightly-coupled approach based on sliding window optimization, where IMU and wheel encoder measurements are fused at the pre-integration stage.
\subsection{Frames and Notations}
The coordinate frames of the sensors include the camera frame, the IMU frame and the odometer frame. The wheel encoder is installed on one rear wheel that always points forward. For the details of these frames the reader may refer to \cite{liu2019visual}. We use $(\cdot)^{w}$ to denote the world frame that is fixed since initialization, and $(\cdot)^{c_k}$, $(\cdot)^{b_k}$ and $(\cdot)^{o_k}$ to denote the camera frame, IMU frame, and odometer frame corresponding to the $k^{th}$ image. Let $\mathbf{R}_A^B$ denote the rotation matrix that takes a vector in frame $\{A\}$ to frame $\{B\}$, and $\mathbf{q}_A^B$ is its quaternion form. $\mathbf{p}_A^B$ is the coordinate of the origin point of frame $\{A\}$ in frame $\{B\}$, and $\mathbf{v}_A^B$ is the velocity of the origin point of frame $\{A\}$ measured in frame $\{B\}$. And let $\mathbf{b}_{a_k}$ and $\mathbf{b}_{\omega_k}$ denote the accelerometer bias and gyroscope bias corresponding to image $k$
respectively. Moreover, we use $[\cdot]_{\times}$ to denote the skew symmetric matrix corresponding to a vector.
\subsection{State Estimation}
\label{ssec:state_estimation}
The parameters to be estimated can be written as
\begin{equation}
\label{eqn1}
\small
\begin{split}
	\mathbf{x} &= \begin{bmatrix} \mathbf{x}_0,\mathbf{x}_1,\hdots\mathbf{x}_{K-1},\lambda_0,\lambda_1,\hdots\lambda_{m-1},\mathbf{R}^b_c,\mathbf{p}^b_c, \mathbf{R}^b_o,\mathbf{p}^b_o \end{bmatrix},\\
	\mathbf{x}_k &= \begin{bmatrix} \mathbf{p}_{b_k}^w, \mathbf{v}_{b_k}^w, \mathbf{q}_{b_k}^w, \mathbf{b}_{a_k}, \mathbf{b}_{\omega_k}\end{bmatrix}, k=0 \hdots K-1,
	\end{split}
\end{equation}
where $\lambda$ is the inverse depth of one landmark in camera frame, $\mathbf{R}^b_c$ and $\mathbf{p}^b_c$ are the camera-IMU extrinsic parameters, while $\mathbf{R}^b_o$ and $\mathbf{p}^b_o$ are the IMU-odometer extrinsic parameters. $m$ is the number of landmarks and $K$ is the size of sliding window.

The cost function $c(\mathbf{x})$ mainly comprises reprojection error terms, IMU-odometer error terms and the marginalization error term, which writes as
\begin{equation}
\label{eqn2}
\small
c(\mathbf{x}) = \sum_L\sum_{j\in \mathcal{B}_L}{\mathbf{e}^v_{L,j}}^\mathsf{T}\mathbf{W}^v\mathbf{e}^v_{L,j} + \sum_{k=0}^{K-2}{\mathbf{e}^s_k}^\mathsf{T}\bm{\Sigma}_{k,k+1}^{-1}{\mathbf{e}^s_k}+{\mathbf{e}^m}^\mathsf{T}\mathbf{e}^m,
\end{equation}
where $\mathbf{e}^v_{L,j}$ means the reprojection residual of landmark $L$ on image $j$, and $\mathbf{W}^v$ is the uniform information matrix for all reprojection error terms. $\mathcal{B}_L$ is the set of images on which landmark $L$ appears. $\mathbf{e}_k^s$ and $\bm{\Sigma}_{k,k+1}$ are the residual vector and covariance matrix of the IMU-odometer terms respectively, which are derived utilizing the IMU-odometer pre-integration results. ${\mathbf{e}^m}^\mathsf{T}\mathbf{e}^m$ is the marginalization error term. In practice we additionally add a very small term confining the roll angle in $\mathbf{R}^b_o$, which is always an unobservable angle because the velocity of the wheel always points forward in its own coordinate frame. This term is so small that it is neglected in the following demonstrations. The nonlinear optimization is performed using Dogleg method by Ceres Solver \cite{ceres-solver}. For the details of state estimation, the reader may refer to \cite{liu2019visual}.

\section{OBSERVABILITY ANALYSIS}
\label{sec:observability}

In this section, we analyze the observability of extrinsic parameters when the platform moves along a straight line with no rotation, which is often the case for a car on a straight road before its first turning. For the observability analysis we need to consider the reprojection constraints
\begin{equation}
\label{eqn3}
\small
\begin{split}
&\pi_c(\mathbf{R}^c_b(\mathbf{R}^{b_j}_w(\mathbf{R}^w_{b_i}(\mathbf{R}^b_c \frac{1}{\lambda_L} \pi_c^{-1}(\begin{bmatrix}\hat{u}_{L, i}\\\hat{v}_{L, i} \end{bmatrix})+\mathbf{p}^b_c)+\mathbf{p}^w_{b_i}-\mathbf{p}^w_{b_j})-\mathbf{p}^b_c))\\&=\begin{bmatrix}\hat{u}_{L, j}\\\hat{v}_{L, j} \end{bmatrix},i=0\hdots K-1, L\in\mathcal{F}_i,j\in\mathcal{V}_l,
\end{split}
\end{equation}
as well as the IMU-odometer constraints
{\small
\begin{align}
\label{eqn4}
&\mathbf{R}_w^{b_k}(\mathbf{p}_{b_{k+1}}^w-\mathbf{p}_{b_k}^w+\frac12 \mathbf{g}^w\Delta t_k^2-\mathbf{v}_{b_k}^w\Delta t_k)-\bm{\alpha}^{b_k}_{b_{k+1}}=\mathbf{0},\\
\label{eqn5}
&\mathbf{R}_w^{b_k}(\mathbf{v}_{b_{k+1}}^w+\mathbf{g}^w\Delta t_k - \mathbf{v}_{b_k}^w)-\bm{\beta}_{b_{k+1}}^{b_k}=\mathbf{0},\\
\label{eqn6}
&2\begin{bmatrix} {(\bm{\gamma}_{b_{k+1}}^{b_k})^{-1} \otimes \mathbf{q}_{b_k}^w}^{-1} \otimes \mathbf{q}_{b_{k+1}}^w \end{bmatrix}_{vec}=\mathbf{0},\\
\label{eqn7}
&\mathbf{R}_w^{b_k}(\mathbf{p}_{b_{k+1}}^w-\mathbf{p}_{b_k}^w)-\mathbf{p}^b_o + \mathbf{R}_w^{b_k}\mathbf{R}_{b_{k+1}}^w\mathbf{p}^b_o-\bm{\eta}_{b_{k+1}}^{b_k}=\mathbf{0},\\
\label{eqn8}
&\mathbf{b}_{a_{k+1}}-\mathbf{b}_{a_k}=\mathbf{0},\\
\label{eqn9}
&\mathbf{b}_{\omega_{k+1}}-\mathbf{b}_{\omega_k}=\mathbf{0}, k=0\hdots K-2,
\end{align}
}%
where $[\hat{u}_{L, i},\hat{v}_{L, i}]^\mathsf{T}$ and $[\hat{u}_{L, j},\hat{v}_{L, j}]^\mathsf{T}$ are the observations of landmark $L$ on image $i$ and image $j$ respectively. $\lambda_L$ is the inverse depth of landmark $L$ in the camera frame where its first observation happens. $\pi_c(\cdot)$ is the projection function, and $\pi_c^{-1}(\cdot)$ is the inverse function of $\pi_c(\cdot)$. $K$ is the size of sliding window, $\mathcal{F}_i$ is the set of landmarks whose first observation happens on image $i$, and $\mathcal{V}_L$ is the set of images that can see landmark $L$ but are not the first image seeing $L$. For more details of (\ref{eqn3}), the reader may refer to \cite{qin2017vins}. $\small \bm{\alpha}^{b_k}_{b_{k+1}}$, $\small \bm{\beta}^{b_k}_{b_{k+1}}$, $\small \bm{\gamma}^{b_k}_{b_{k+1}}$ and $\small \bm{\eta}^{b_k}_{b_{k+1}}$ are the nominal states from IMU-odometer pre-integration, after being compensated by the changes in estimated IMU biases and IMU-odometer extrinsic parameters. Among them $\small \bm{\gamma}^{b_k}_{b_{k+1}}$ is the rotation from the IMU frame of image $k+1$ to that of image $k$. $\small \bm{\eta}^{b_k}_{b_{k+1}}$ is the displacement between the origin of odometer frame of image $k$ and that of image $k+1$, integrated using gyroscope and odometer measurements. $\small \bm{\alpha}^{b_k}_{b_{k+1}}$ and $\small \bm{\beta}^{b_k}_{b_{k+1}}$ do not have straightforward geometrical meanings, but they have the same dimensions as displacement and velocity  respectively. The readers may look at (3) and (5) in \cite{qin2017vins} for details of $\small \bm{\alpha}^{b_k}_{b_{k+1}}$, $\small \bm{\beta}^{b_k}_{b_{k+1}}$ and $\small \bm{\gamma}^{b_k}_{b_{k+1}}$, as well as (1) and (12) in \cite{liu2019visual} for details of $\bm{\eta}^{b_k}_{b_{k+1}}$ respectively. $\Delta t_k$ is the time interval between image $k$ and image $k+1$. And $[\cdot]_{vec}$ denotes the vector part of a quaternion. 

Among the extrinsic parameters $\small (\mathbf{R}^b_c,\mathbf{p}^b_c)$ and $\small (\mathbf{R}^b_o,\mathbf{p}^b_o)$, $\small \mathbf{p}^b_o$ is only involved in (\ref{eqn7}), which can be rewritten as
\begin{equation}
\small
\label{eqn10}
\mathbf{R}_w^{b_k}(\mathbf{p}_{b_{k+1}}^w-\mathbf{p}_{b_k}^w) + (\mathbf{R}_w^{b_k}\mathbf{R}_{b_{k+1}}^w-\mathbf{I})\mathbf{p}^b_o-\bm{\eta}_{b_{k+1}}^{b_k}=\mathbf{0}.
\end{equation}
When the platform moves along a straight line with no rotation, $\small \mathbf{R}_w^{b_k}\mathbf{R}_{b_{k+1}}^w-\mathbf{I}=\mathbf{O}$, in which case $\small \mathbf{p}^b_o$ is not involved in any of the constraints, so it is unobservable.

Similarly, $\small \mathbf{R}^b_c$ and $\small \mathbf{p}^b_c$ are only involved in (\ref{eqn3}), which can be rewritten as
\begin{equation}
\label{eqn11}
\small
\begin{split}
&\pi_c(\frac{1}{\lambda_L}\mathbf{R}^c_b\mathbf{R}^{b_j}_w\mathbf{R}^w_{b_i}\mathbf{R}^b_c \pi_c^{-1}(\begin{bmatrix}\hat{u}_{L, i}\\\hat{v}_{L,i} \end{bmatrix})+\mathbf{R}^c_b(\mathbf{R}^{b_j}_w\mathbf{R}^w_{b_i}-\mathbf{I})\mathbf{p}^b_c\\&+\mathbf{R}^c_b\mathbf{R}^{b_j}_w(\mathbf{p}^w_{b_i}-\mathbf{p}^w_{b_j}))=\begin{bmatrix}\hat{u}_{L, j}\\\hat{v}_{L, j} \end{bmatrix},i=0\hdots K-1, L\in\mathcal{F}_i,j\in\mathcal{V}_l.
\end{split}
\end{equation}
When the platform moves along a straight line with no rotation, $\small \mathbf{p}^b_c$ is unobservable because $\small \mathbf{R}_w^{b_j}\mathbf{R}_{b_i}^w-\mathbf{I}=\mathbf{O}$. Moreover, in such a case, for every image pair $(i,j)$, we have $\small \mathbf{R}^{b_j}_w(\mathbf{p}^w_{b_i}-\mathbf{p}^w_{b_j}) = s_{i,j}\mathbf{d}$, where $s_{i,j}$ is a scalar, and $\mathbf{d}$ is a unit vector denoting the driving direction, which is independent of $(i, j)$. In such a case, (\ref{eqn11}) can be rewritten as
\begin{equation}
\label{eqn18}
\small
\pi_c(\frac{1}{\lambda_L} \pi_c^{-1}(\begin{bmatrix}\hat{u}_{L, i}\\\hat{v}_{L,i} \end{bmatrix})+s_{i,j}\mathbf{R}^c_b\mathbf{d})=\begin{bmatrix}\hat{u}_{L, j}\\\hat{v}_{L, j} \end{bmatrix},i=0\hdots K-1, L\in\mathcal{F}_i,j\in\mathcal{V}_l,
\end{equation}
 From (\ref{eqn18}) it is clear that the component in rotation $\small \mathbf{R}^b_c$ corresponding to the rotation around the driving direction $\mathbf{d}$, which is also called the roll angle in the following, is unobservable as well.

\begin{figure}[htb]
\centering
\includegraphics[width=6.5cm]{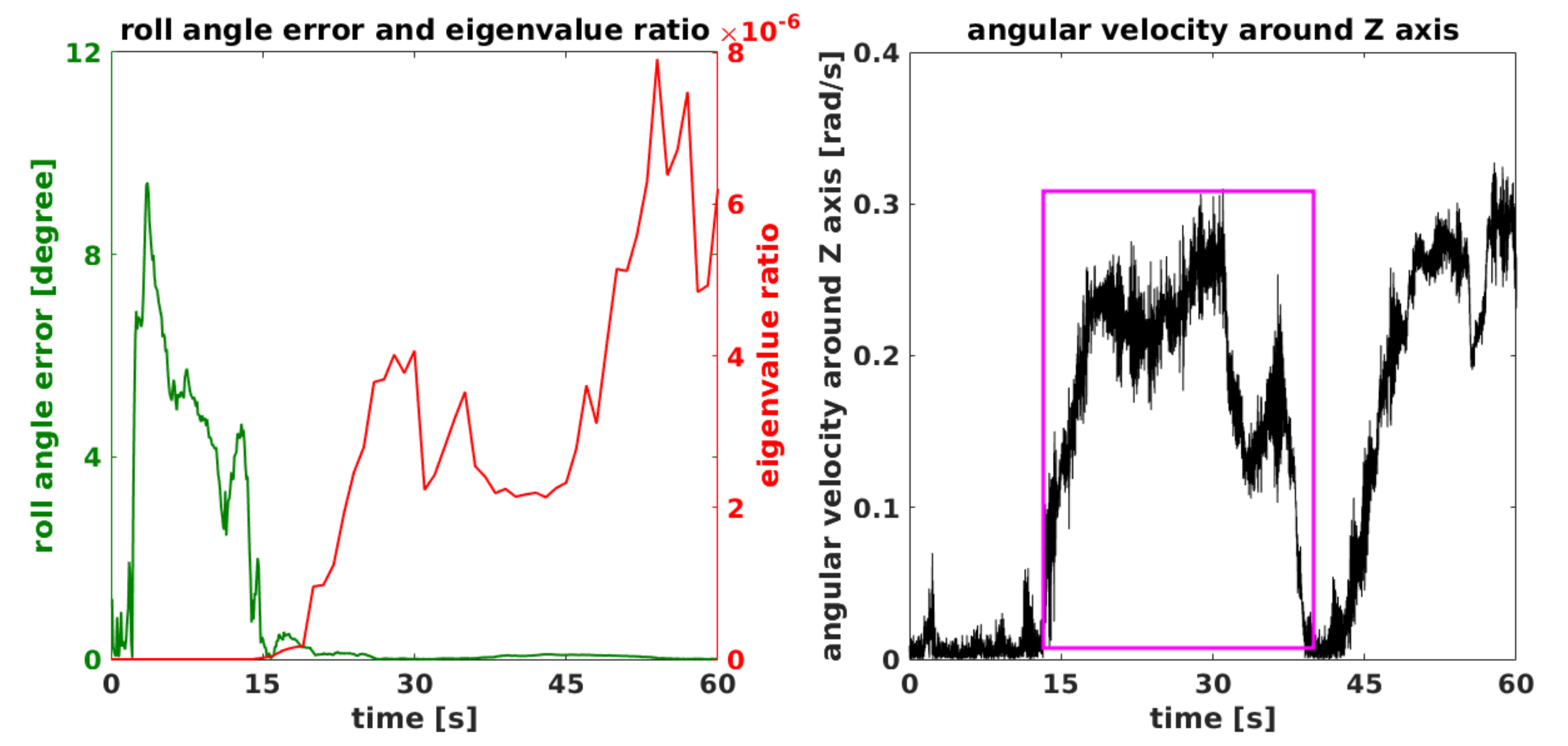}
\caption{Estimation error in roll angle of $\mathbf{R}^b_c$, the eigenvalue ratio and the angular velocity around the Z axis (the vertical axis) at the first turning in sequence urban34. The pinkish box indicates the first turning. The system starts at a dozen of seconds before the first turning.}
\label{fig_eigenratio}
\end{figure}

In practice, among the seven unobservable directions (three in $\mathbf{p}^b_o$, three in $\mathbf{p}^b_c$ and one in $\mathbf{R}^b_c$) when the platform moves along a straight line with no rotation, the roll angle in $\mathbf{R}^b_c$ is the direction whose observability is most relevant to whether the platform has made a turn. We firstly perform eigendecomposition of the Hessian matrix, and then compute the ratio of the eigenvalue corresponding to the eigenvector which is most approximate to direction of the perturbation of the roll angle in $\mathbf{R}^b_c$, to the largest eigenvalue.  For the experiment illustrated in Figure \ref{fig_eigenratio}, the accelerometer bias is held constant and the extrinsic parameters start to be adjusted since the beginning. It is clear from Figure \ref{fig_eigenratio} that after the first turning the error in roll angle of $\mathbf{R}^b_c$ dramatically decreases and that the eigenvalue ratio dramatically increases, both of which indicate that the roll angle in $\mathbf{R}^b_c$ can be estimated much better after the first turning. Compared with  \cite{schneider2019observability}, our observability analysis computes the eigenvalue ratio inside sliding window that keeps sliding, rather than fisher information matrix inside previously divided segments, thereby illustrating the relationship between turning and the observability of extrinsic parameters more clearly. Moreover, our analysis reveals that the extrinsic parameter is still observable after the turning given that marginalization is applied, because the eigenvalue ratio does not dramatically decrease after the turning.

Practically, when the motion of the platform is approximately but not exactly along a straight line, the errors in the unobservable directions of the extrinsic parameters do affect the accuracy of the trajectory. Table \ref{table_eps} shows the comparison of absolute trajectory error (ATE) for 4 sequences in KAIST Urban Data Set\cite{jeong2019complex}, either using the accurate extrinsic parameters calibrated offline or using the extrinsic parameters with added fixed error (5 degrees in the roll angle of $\mathbf{R}^b_c$). More details of observability analysis are in our arXiv technical report.

\begin{table}[htb]
\renewcommand{\arraystretch}{1.3}
\scriptsize
\caption{ATE (in meters) using different extrinsic parameters}
\label{table_eps}
\centering
\begin{threeparttable}
\begin{tabular}{*{5}{c}}
\bottomrule
Sequence (urban-)&22 (3.4km)&23 (3.4km)&24 (4.2km)&25 (2.5km)\\
\midrule
EPs accurate&\textbf{8.8}&\textbf{11.1}&\textbf{15.0}&\textbf{8.0}\\
EPs with error&14.2&13.5&16.2&9.7\\
\bottomrule
\end{tabular}
\begin{tablenotes}
        \fontsize{8pt}{10pt}
		\item Here \emph{\scriptsize EPs accurate} means using the accurate extrinsic parameters calibrated offline, and \emph{\scriptsize EPs with error} means using the extrinsic parameters with added fixed error. \emph{\scriptsize ATE} means absolute trajectory error, and \emph{\scriptsize EPs} mean extrinsic parameters. In each of the four sequences the car moves along an approximately straight road, and the trajectories are computed using the state estimation method illustrated in Section \ref{ssec:state_estimation}, with accelerometer bias and extrinsic parameters both held constant. 
\end{tablenotes}
\end{threeparttable}
\end{table}

\section{METHOD}
\label{sec:method}
Taking into consideration that for the odometer-aided VI-SLAM system described in Section \ref{ssec:state_estimation}, the system is not stable and the extrinsic parameters can not be correctly estimated in the beginning, and that the accelerometer bias is unobservable until the platform makes a turn, we propose a robust method to acquire accurate real-time trajectory.
\subsection{Forward Computation Thread and Backward Computation Thread}
\label{ssec:two_threads}
\begin{figure}[htb]
\centering

  \includegraphics[width=4.6cm]{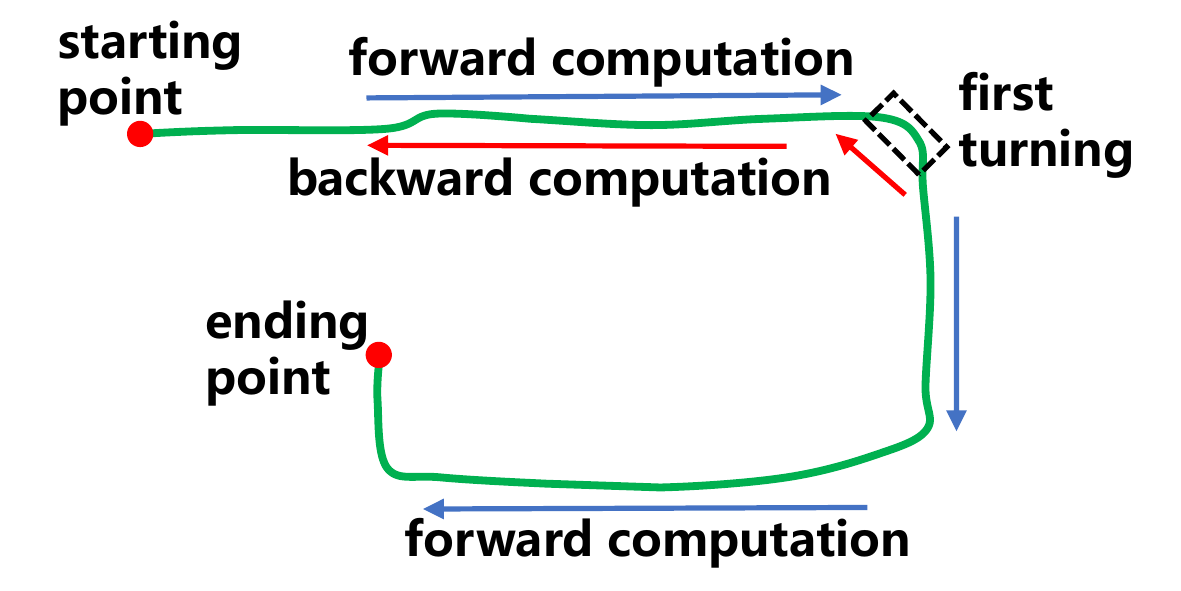}
  
\caption{Schematic diagram about forward computation and backward computation. Forward computation starts from the beginning. After the first turning backward computation starts. Forward computation continues to operate until the end. Backward computation proceeds to the starting point to work out more accurate poses.}
\label{fig_schematic}
\end{figure}

In the very beginning, we propagate the poses and try to initialize our system. After the system is initialized as described in \cite{liu2019visual}, the state estimation in sliding window is performed as in Section \ref{ssec:state_estimation} in the main thread, which we call the forward computation thread. In the forward computation thread, before the first turning, the extrinsic parameters are held constant, and the accelerometer bias is set to zero and held constant, in order to make the system robust in the beginning. At this stage, we limit the magnitude of the marginalization term as described in Section \ref{ssec:bounded_marginalization}. Once the platform has made a turn, the accelerometer bias starts to be adjusted, and as soon as the estimation of  accelerometer bias reaches convergence according to the criterion adopted in \cite{liu2019visual}, the extrinsic parameters starts to be adjusted. The detection criteria for a turning will be described and explained in Sect.\ref{ssec:turning_detection}. Thanks to the fact that the marginalization term contains historical information, especially the information gathered during the first turning, the accelerometer bias and extrinsic parameters that are engaged in state estimation will soon reach their desired values. After a time interval $T_2$ (30 seconds in our experiments) since the extrinsic parameters begin to be adjusted, we create a new thread named backward computation thread, meanwhile the forward computation thread keeps running. Both the two threads are independent with each other. Figure \ref{fig_schematic} is the schematic diagram illustrating forward computation and backward computation.

The backward computation thread also performs state estimation in a sliding window where the parameters are as (\ref{eqn1}) and the cost function writes as (\ref{eqn2}). When creating the backward computation thread, the values of the parameters in the backward computation thread except for landmark inverse depths are copied from the forward computation thread, and the IMU-odometer terms and the marginalization term in the backward computation thread are identical to their counterparts in the forward computation thread. For the backward computation thread, the reprojection errors still take the form of (\ref{eqn3}), while the \emph{first observation} in (\ref{eqn3}) means the observation happening on the image with the latest timestamp, instead of the one with the earliest timestamp as in the forward computation thread. Meanwhile, the inverse depth of each landmark $L$ is shifted as
\begin{equation}
\small
\lambda_L = 1/(\mathbf{e}_3^\mathsf{T}\mathbf{R}^c_b(\mathbf{R}^{b_q}_w(\mathbf{R}^w_{b_p}(\mathbf{R}^b_c \frac{1}{\lambda'_L} \pi_c^{-1}(\begin{bmatrix}\hat{u}_{L, p}\\\hat{v}_{L, p} \end{bmatrix})+\mathbf{p}^b_c)+\mathbf{p}^w_{b_p}-\mathbf{p}^w_{b_q})-\mathbf{p}^b_c)),
\end{equation}
where $\lambda_L$ and $\lambda'_L$ are the inverse depths of landmark $L$ after and before shifting respectively, $\mathbf{e}_3^\mathsf{T}=\begin{bmatrix}0&0&1\end{bmatrix}$, $p$ and $q$ are indexes of the earliest and latest image which can see the landmark $L$ in the sliding window respectively, and the meanings of the other symbols are the same as those in (\ref{eqn3}).
\begin{figure}[htb]
\centering
\subfigure[sliding window in the forward computation thread]{
\includegraphics[width=6.2cm]{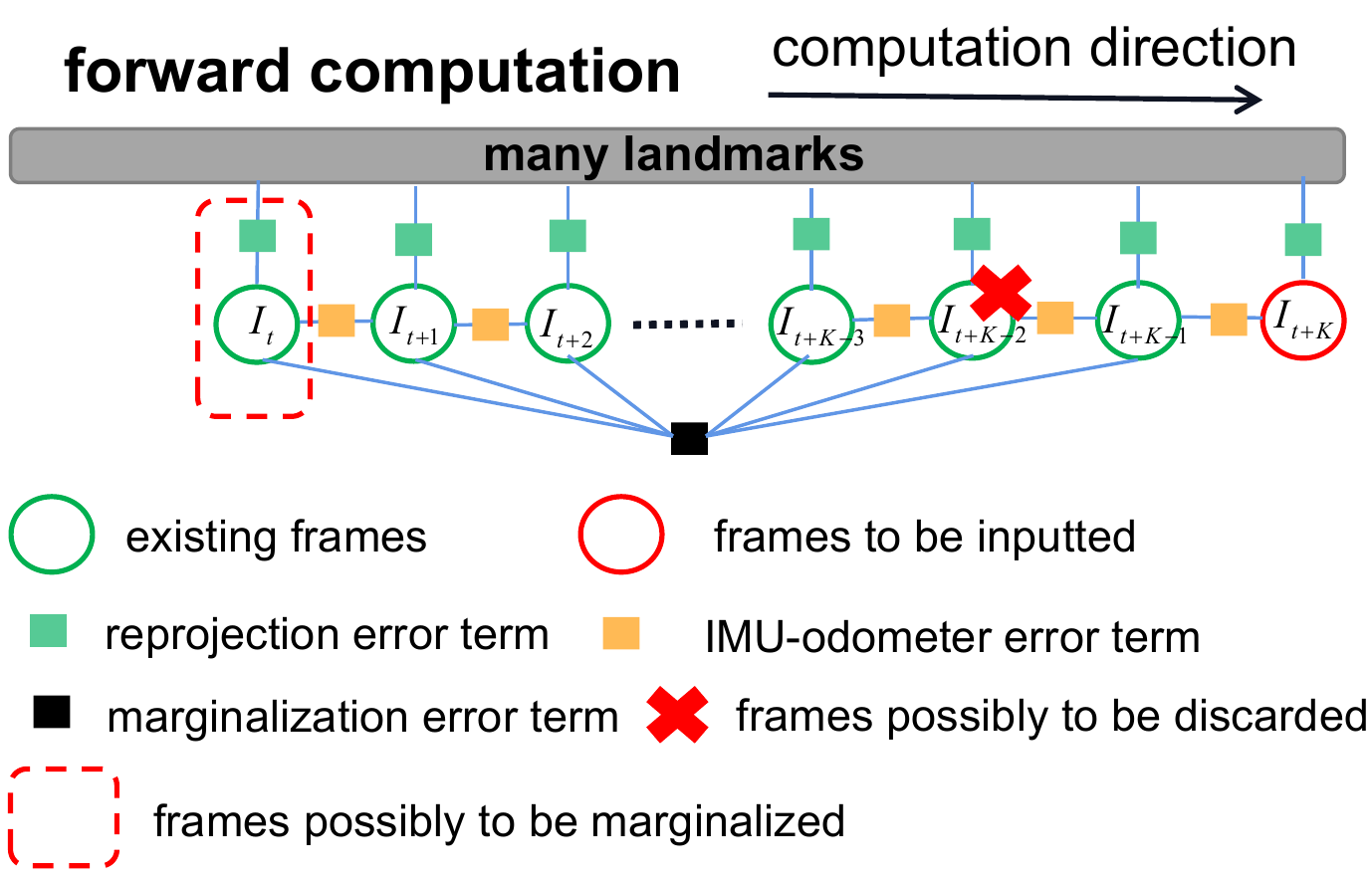}
}
\subfigure[sliding window in the backward computation thread]{
\includegraphics[width=6.2cm]{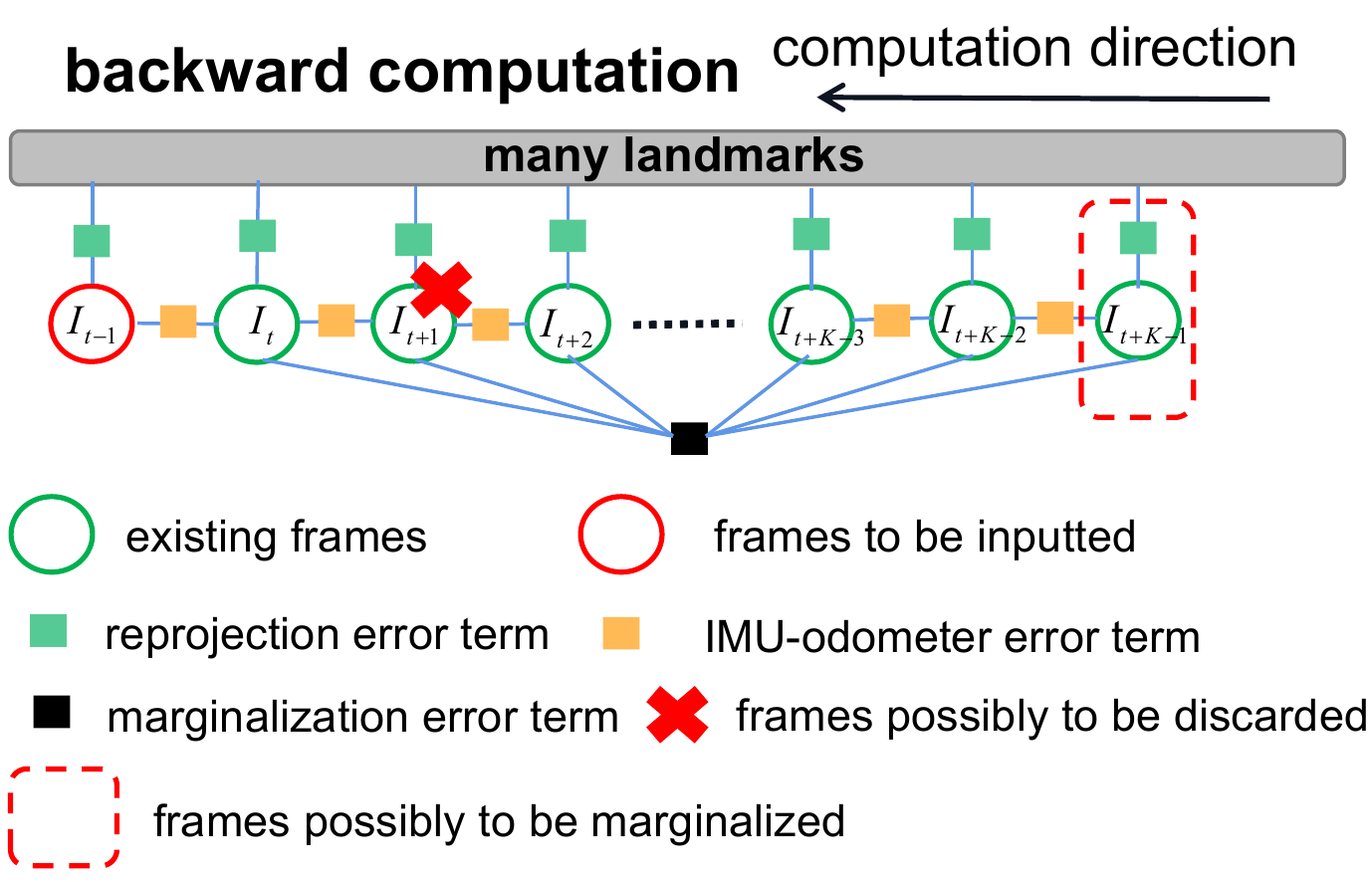}
}
\caption{Contrast between the sliding windows in forward and backward computation threads. In either of the two threads, when a new frame is inputted, the frame marked with red cross is discarded if it is not a keyframe. Otherwise, the frame in the dashed red box is marginalized.}
\label{fig_contrast_backward}
\end{figure}

Parameters can be estimated correctly in the backward computation thread since the backward computation thread starts, because of the information contained in the marginalization error term. Figure \ref{fig_contrast_backward}(a)-\ref{fig_contrast_backward}(b) show the contrast between the sliding windows in forward and backward computation threads. In the following we illustrate how the backward computation thread operates according to Figure \ref{fig_contrast_backward}(b). Suppose that the first frame in data sequence is $I_0$, and that at a certain moment $t$, there are $K$ frames in the sliding window, namely $I_t, I_{t+1}\hdots I_{t+K-1}$. In the backward computation thread, every time the nonlinear optimization in the sliding window finishes at the certain moment $t$, the next frame to be inputted is $I_{t-1}$, which is the one previous to the frame $I_t$ whose timestamp is the earliest in the sliding window. The IMU-odometer pre-integration between the above two frames ($I_{t-1}$ and $I_t$) is computed, and the initial value of the pose and velocity of the frame to be newly inputted ($I_{t-1}$) is propagated using IMU measurements between the two frames. Note that although the IMU-odometer pre-integration between the above two frames has been performed in the past in the forward computation thread, recomputation is needed because the estimated value of IMU biases and the extrinsic parameter $\mathbf{R}^b_o$ have changed, and they are engaged in pre-integration. Next, if the frame with the second earliest timestamp ($I_{t+1}$) is not a keyframe, it is discarded. Otherwise, the frame with the latest timestamp ($I_{t+K-1}$) is marginalized. The criterion to judge whether an image frame is a keyframe is the same as that in \cite{qin2017vins}. The backward computation terminates when the first frame in data sequence $I_0$ has been inputted into the sliding window. The IMU and wheel encoder measurements and the feature points used in backward computation are recorded previously during the forward computation. When the pose of a certain frame is estimated in the backward computation thread, it is used to substitute the corresponding pose previously estimated in the forward computation thread, because the poses estimated in the backward computation thread are more accurate.

\subsection{Turning Detection}
\label{ssec:turning_detection}
\begin{figure}[htb]
\centering
\subfigure[]{
\includegraphics[width=4.0cm]{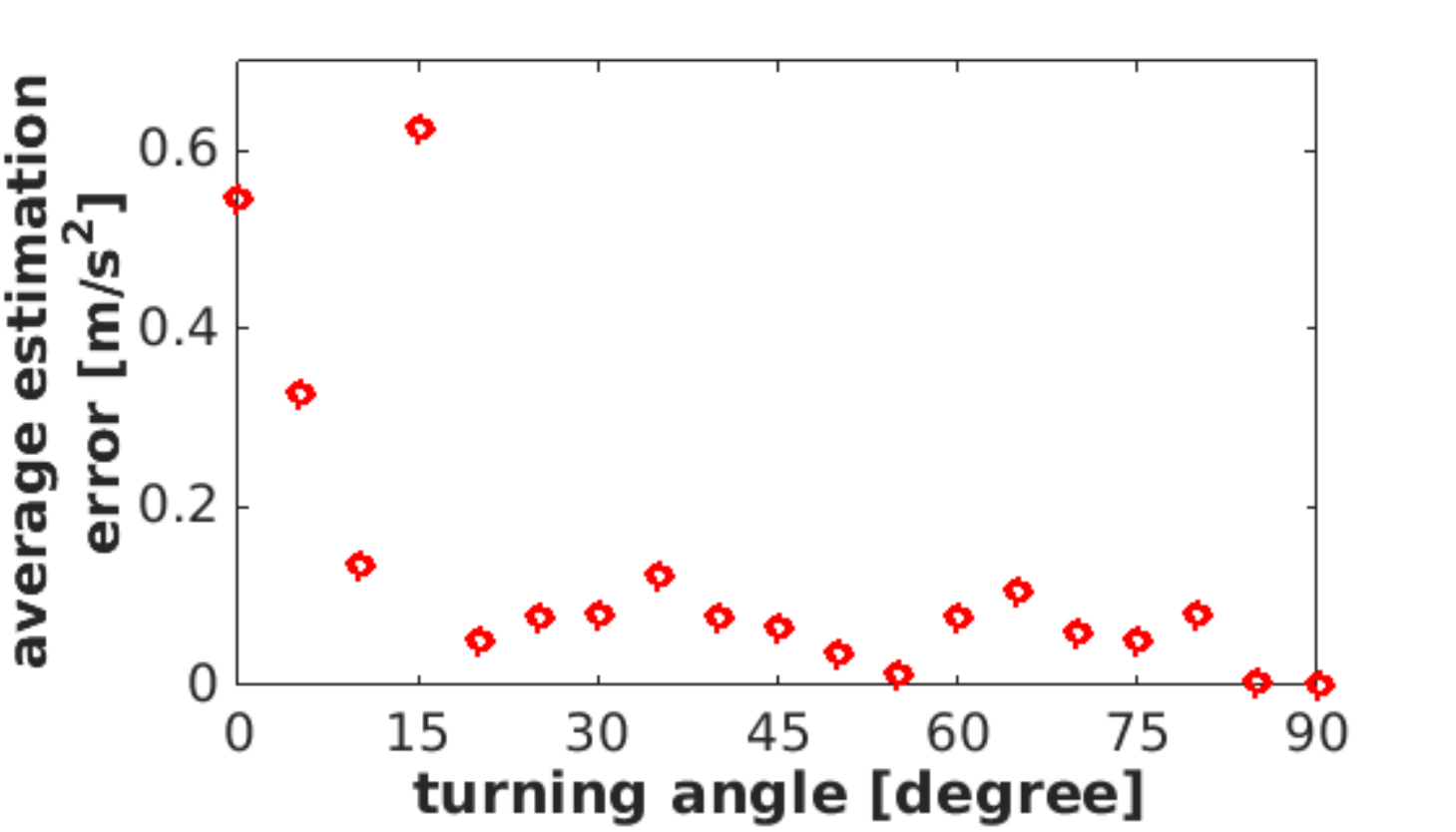}
}
\subfigure[]{
\includegraphics[width=4.0cm]{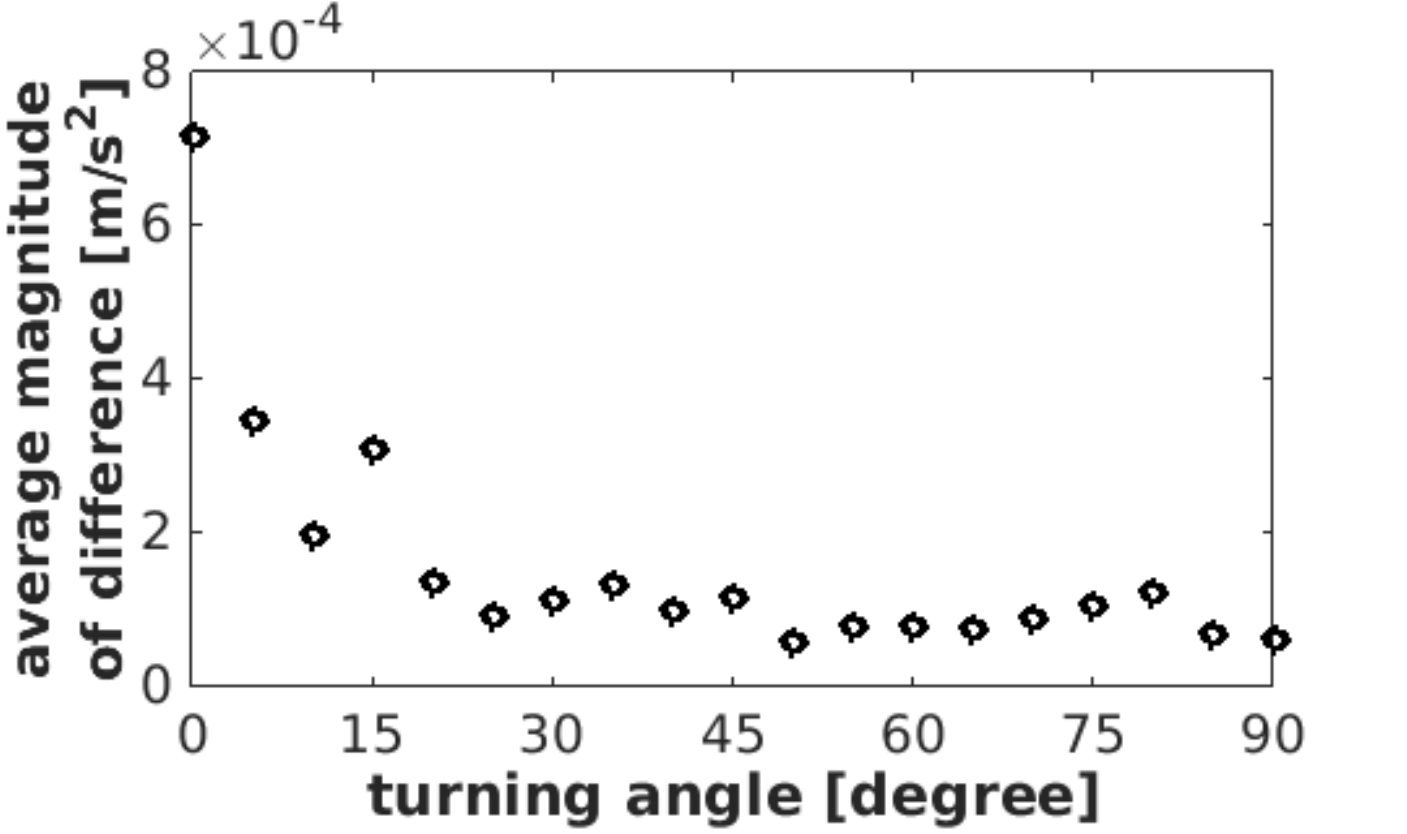}
}
\caption{(a) average estimation error in accelerometer biases w.r.t. ground-truth accelerometer biases after turning and (b) average magnitude of difference between every two successively estimated accelerometer biases after turning. Each dot in the figures represents the average value in a data sequence that contains a certain turning angle. The difference between every two successively estimated accelerometer biases is also expected to be small, because the accelerometer bias is a slow time-varying quantity.}
\label{fig_simulate}
\end{figure}

Since the local optimizations are always performed inside a sliding window, for turning detection we consider the turning angle inside a sliding window. We determine how large a turning angle should reach to result in good estimation of accelerometer bias, because the estimation of accelerometer bias is very relevant to the turning of the platform, which is evident from Figure \ref{fig_comp} and \cite{liu2019visual}. We simulate 19 data sequences, each of which contains a single turning that can be contained inside a sliding window, and the turnings vary every 5 degrees from 0 to 90 degrees. We run the odometer-aided visual-inertial SLAM with only forward computation on each sequence, and the average estimation error in accelerometer biases after turning, as well as the average magnitude of difference between every two successively estimated accelerometer biases after turning are shown in Figure \ref{fig_simulate}(a)-\ref{fig_simulate}(b).

From Figure \ref{fig_simulate}(a)-\ref{fig_simulate}(b) we can see that when the turning angle exceeds 20 degrees, both the average estimation error in accelerometer biases after turning and the average magnitude of difference between every two successively estimated accelerometer biases after turning do not further decrease as the turning angle increases, indicating that the estimation of accelerometer bias is good enough under such circumstances. Therefore, we consider the platform has made a turn if and only if it has turned larger than 20 degrees within a sliding window. We do not use eigenvalue ratio to detect turning, as eigendecomposition is time-consuming. On the data sequence urban25, eigendecomposition takes $10.5ms$ on average after each optimization, and updating the turning angle takes $0.0007ms$ on average between two consecutive optimizations.
\subsection{Computation of Real-time Trajectory}
\begin{figure}[htb]
  \centering
  \subfigure[Trajectory comprising poses $(\mathbf{R}'^w_{c_k},\mathbf{p}'^w_{c_k})$]{
  \label{subfig_0min}
  \centering
  \includegraphics[width=3.7cm]{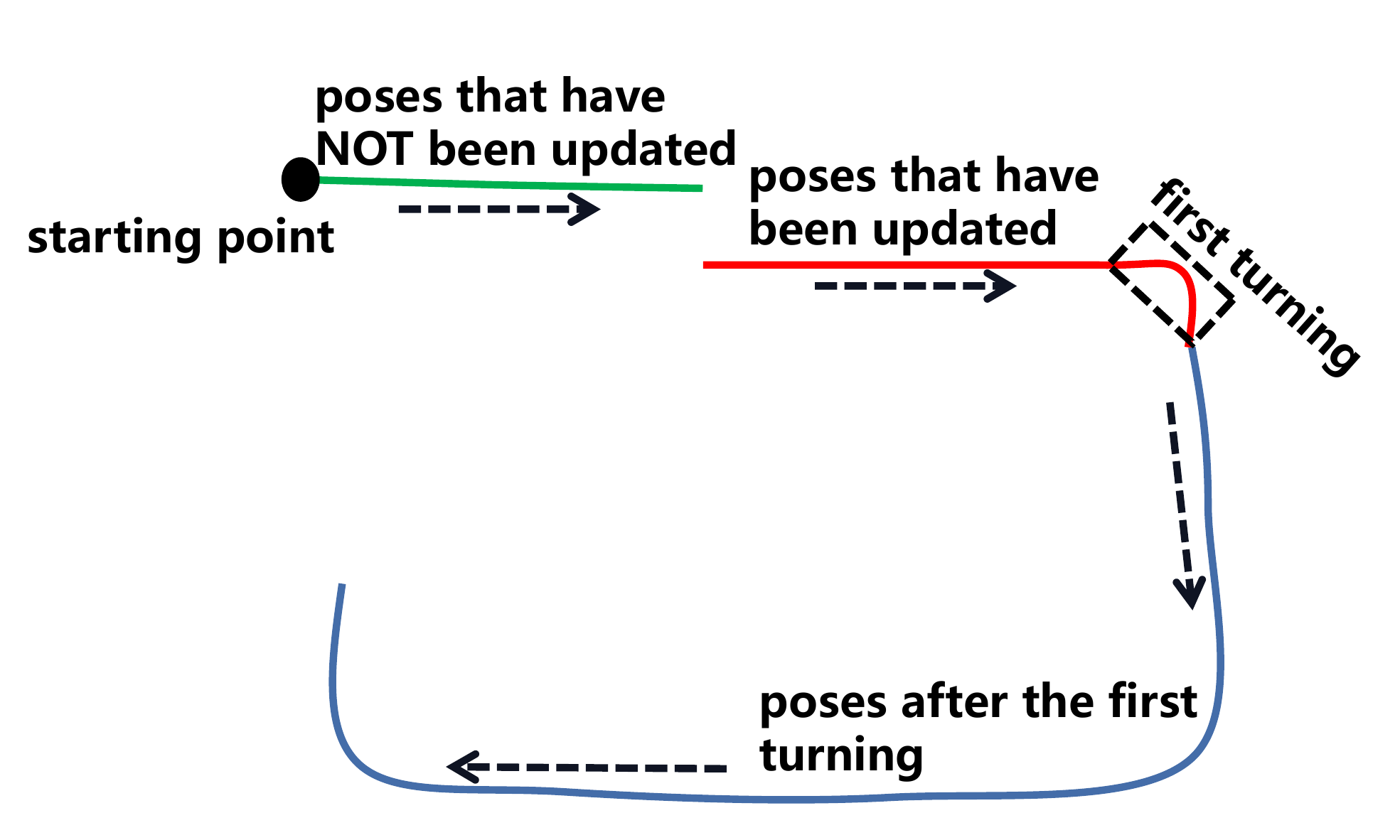}
  }
  \subfigure[Real-time trajectory comprising poses $(\mathbf{R}''^w_{c_k},\mathbf{p}''^w_{c_k})$]{
  \label{subfig_3min}
  \centering
  \includegraphics[width=3.7cm]{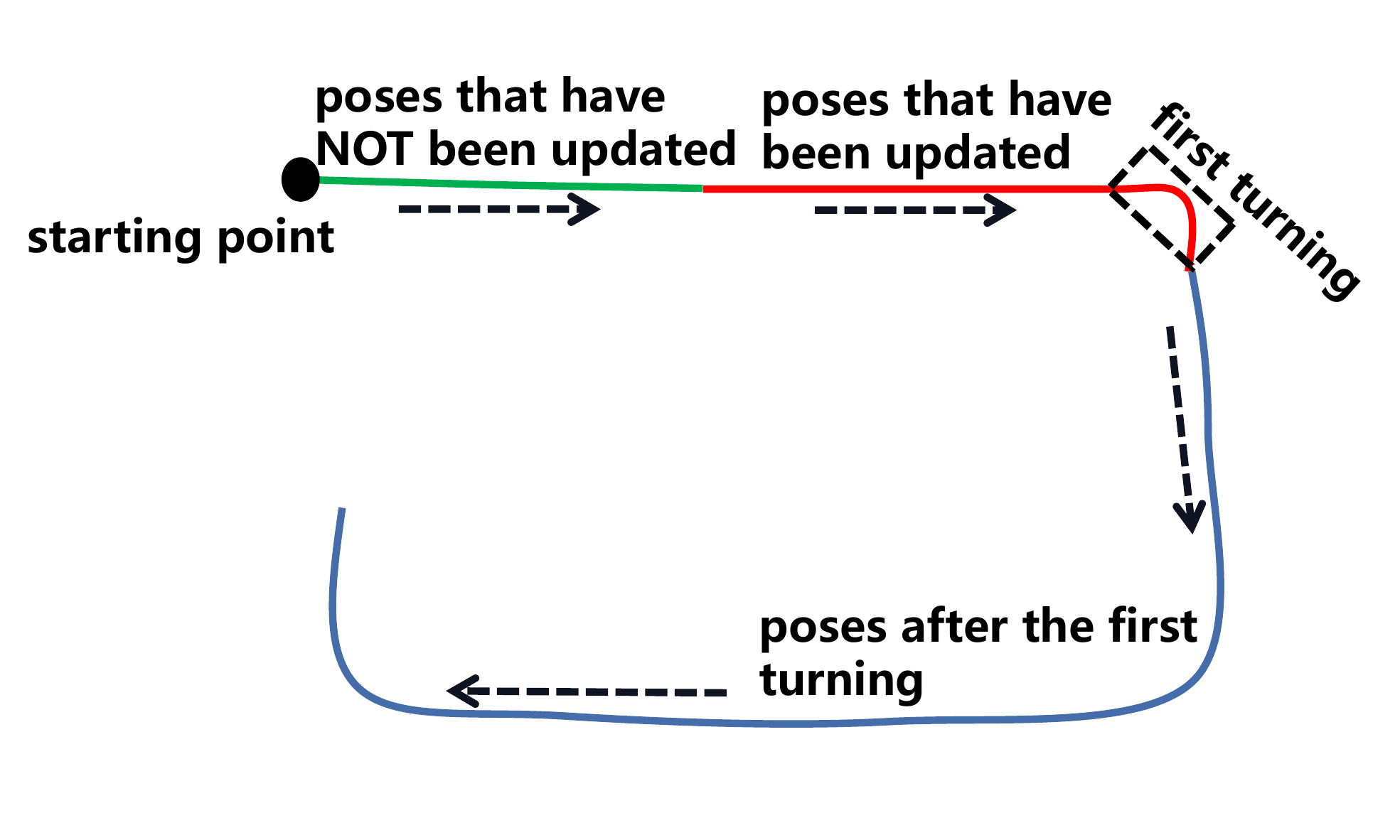}
  }
\caption{Schematic diagram of real-time trajectory computation. The black dotted arrows represent the moving direction of the platform.} 
\label{fig_realtimesk}
\end{figure}
Although the backward computation directly updates the past poses, every time the backward computation thread updates the pose of a certain frame, the current pose is also adjusted accordingly. As can be seen from Figure \ref{fig_realtimesk}, The real-time trajectory is computed by keeping the starting point unchanged and keeping the trajectory continuous. For a certain image frame $k$, its pose $(\mathbf{R}'^w_{c_k},\mathbf{p}'^w_{c_k})$ is first computed by forward computation thread, and if it is recomputed in the backward computation thread, its pose $(\mathbf{R}'^w_{c_k},\mathbf{p}'^w_{c_k})$ is updated by the backward computation thread, and in this case the pose previously computed by the forward computation thread is denoted as $(\hat{\mathbf{R}}^w_{c_k},\hat{\mathbf{p}}^w_{c_k})$. Let $j$ denote the frame that has just been updated by the backward computation thread. For the frames before frame $j$, i.e. $i<j$, the real-time pose $({\mathbf{R}''}^w_{c_i},{\mathbf{p}''}^w_{c_i})$ equals $({\mathbf{R}'}^w_{c_i},{\mathbf{p}'}^w_{c_i})$, that is to say they are unchanged. For the frames after frame $j$, i.e. $i\geq j$, the real-time pose $({\mathbf{R}''}^w_{c_i},{\mathbf{p}''}^w_{c_i})$ is computed as
\begin{equation}
\small
\label{eqn12}
\begin{split}
&{\mathbf{R}''}^w_{c_i} = {\hat{\mathbf{R}}}^w_{c_j}{\mathbf{R}'}_w^{c_j}{\mathbf{R}'}^w_{c_i},\\
&{\mathbf{p}''}^w_{c_i} = {\hat{\mathbf{R}}}^w_{c_j}{\mathbf{R}'}_w^{c_j}({\mathbf{p}'}^w_{c_i}-{\mathbf{p}'}^w_{c_j})+\hat{\mathbf{p}}^w_{c_j}, \text{ for } i \geq j.
\end{split}
\end{equation}

\subsection{Bounded Marginalization Term}
\label{ssec:bounded_marginalization}
The marginalization residual takes the form of $\mathbf{e}^m=\mathbf{r}^m-\mathbf{J}^m\delta\mathbf{x}$. $\mathbf{r}^m$ and $\mathbf{J}^m$ are computed in the marginalization process, and $\delta\mathbf{x}$ is the step to update the parameters $\mathbf{x}$. We have observed the phenomenon that the marginalization error keeps growing and thus the total error keeps growing before the first turning, when the accelerometer bias and extrinsic parameters are held constant. 
In order to reduce the accumulation of the error caused by inaccurate extrinsic parameters and accelerometer bias in the marginalization error term and prevent the above error from dominating the state estimation, before the first turning we multiply $\mathbf{r}^m$ and $\mathbf{J}^m$ by a ratio $\mu$ once the ratio of marginalization error ${\mathbf{e}^m}^\mathsf{T}{\mathbf{e}^m}$ to total error $c(\mathbf{x})$ in (\ref{eqn2}) after optimization rises beyond a threshold $r$. 
In our experiment $\mu$ is set to 0.85 and $r$ is set to 0.4.

\section{EXPERIMENTS}
We evaluate the effect of the bidirectional trajectory computation method proposed in this paper on KAIST Urban Data Set \cite{jeong2019complex}, which is a publicly available  dataset containing data in complex urban scenes collected on a rear wheel drive passenger car. The sensors in the dataset include stereo cameras, one IMU and two wheel encoders mounted on two rear wheels. The frequencies of the captured images, IMU measurements and wheel encoder measurements are 10Hz, 100Hz and 100Hz respectively. The proposed approach is compared with the stereo inertial version of the state-of-the-art VI-SLAM system VINS-Fusion\cite{qin2019a}, the standard VIO \cite{li2013optimization}, and the odometer-aided VI-SLAM approaches \cite{Zhang2019vision}, \cite{zhang2019localization} and \cite{liu2019visual}. The proposed approach and \cite{liu2019visual} use a monocular camera, one IMU and one wheel encoder. The approaches \cite{Zhang2019vision} and \cite{zhang2019localization}, as reported in their papers, use a monocular camera, one IMU and two wheel encoders. The stereo inertial version of VINS-Fusion uses stereo cameras and one IMU. To our knowledge, there are no other approaches intentionally dealing with the unobservability after initialization and before the first turning. All the experiments presented are performed on a PC with Intel Core i7 3.6GHz $\times$ 6 core CPU and 64GB memory. The extrinsic parameters provided in the dataset are adopted as initial values, which may not be very accurate.

\subsection{Average Positioning Error by Aligning the Starting Frame}
\label{ssec:average_positioning_error}
The primary concern of our bidirectional trajectory computation method is to improve the accuracy at initial stage, which matters a lot supposing we only know the position and orientation of the vehicle at the starting point. In our first evaluation, we align the position and orientation of the starting image frames for the resulting trajectory from VI-SLAM approaches and the ground truth trajectory, and compute the average positioning error of every frame in the data sequence. The practice of aligning the starting frames is also adopted in the evaluation criteria on KITTI dataset \cite{Geiger2012CVPR}. Here our proposed approach is mainly compared with \cite{liu2019visual}, which our approach is based on. The work \cite{liu2019visual} starts to optimize accelerometer bias from the beginning, and fix the extrinsic parameters until the platform has made a turn and the estimation of accelerometer bias has reached convergence, in order to reduce the instability in the very beginning. In order to make an exhaustive comparison on different strategies dealing with accelerometer bias and extrinsic parameters that are two instability factors, we derive some adapted versions from \cite{liu2019visual}, that are: (i) both accelerometer bias and extrinsic parameters are fixed until the first turning (FAFE), (ii) the extrinsic parameters starts to be optimized from the beginning, and accelerometer bias is fixed until the first turning (FAOE), (iii) both accelerometer bias and extrinsic parameters starts to be optimized from the beginning (OAOE). Our proposed approach is firstly compared against \cite{liu2019visual} (OAFE) and its three adapted versions in the above, as well as VINS-Fusion\cite{qin2019a}. To make a fair comparison, we select the image frame when the vehicle has traveled 100 meters as the starting frame, to avoid being affected by some erroneous pose estimations from some approaches in the very beginning. This comparison is made on all the 15 sequences with stereo cameras and with complexity level 3 (middle) or level 4 (high) in \cite{jeong2019complex}, namely urban25-urban39. The comparison of average positioning error by aligning the starting frame is shown in Table \ref{table_average_error}.

\begin{table}[!t]
\renewcommand{\arraystretch}{1.3}
\caption{Comparison of average positioning error (in meters) by aligning the starting frame}
\scriptsize
\label{table_average_error}
\centering
\begin{threeparttable}
\begin{tabular}{*{7}{c}}
\bottomrule
\multirow{2}{*}{Sequence}&\multirow{3}{*}{Proposed}&\multirow{3}{*}{FAFE}&\multirow{3}{*}{FAOE}&\multirow{3}{*}{OAOE}&\multirow{2}{*}{OAFE}&VINS-\\
&&&&&&Fusion\\
(urban-)&&&&&\cite{liu2019visual}&\cite{qin2019a}\\
\midrule
*25 (2.5km)&11.6&\textbf{11.3}&72.7&62.1&15.3&862.7\\
26 (4.0km)&\textbf{20.8}&41.0&42.8&29.8&41.7&52.9\\
27 (5.4km)&\textbf{19.0}&49.5&73.4&91.9&44.5&63.1\\
28 (11.5km)&32.6&61.5&47.7&\textbf{27.7}&104.8&103.4\\
29 (3.6km)&51.3&44.3&\textbf{12.1}&13.3&40.1&122.6\\
30 (6.0km)&\textbf{24.8}&34.6&36.8&45.5&43.0&$\times$\\
31 (11.4km)&\textbf{389.9}&1108&995.9&1072&1230&1739\\
32 (7.1km)&\textbf{66.9}&407.5&140.6&149.1&422.6&257.9\\
33 (7.6km)&\textbf{31.8}&177.3&81.9&130.9&221.3&696.7\\
34 (7.8km)&118.9&\textbf{98.8}&160.9&122.6&168.7&$\times$\\
*35 (3.2km)&61.1&\textbf{49.3}&290.9&253.4&57.5&$\times$\\
36 (9.0km)&\textbf{218.0}&333.1&221.5&281.0&283.7&$\times$\\
*37 (11.8km)&462.9&\textbf{371.0}&1989&2221&677.0&1126\\
38 (11.4km)&\textbf{27.6}&123.4&151.9&44.8&101.3&134.6\\
39 (11.0km)&\textbf{13.5}&953.7&22.0&36.8&42.1&$\times$\\
\bottomrule
\end{tabular}
\begin{tablenotes}
        \fontsize{8pt}{10pt}
		\item Here \emph{\scriptsize Proposed} means the proposed approach in this paper and '$\times$' means failure. The \emph{\scriptsize Sequence} column firstly contains the sequence number which is $urban$XX, followed by the trajectory length in the parentheses. The sequences marked with '*' do not contain turnings, so the difference in accuracies on those sequences between the proposed approach and \emph{\scriptsize FAFE} is only resulted by restricting the marginalization error as described in Section \ref{sec:method}. \emph{\scriptsize FAFE} refers to fixing accelerometer bias and fixing extrinsic parameters. \emph{\scriptsize FAOE} refers to fixing accelerometer bias and optimizing extrinsic parameters; \emph{\scriptsize OAOE} refers to optimizing accelerometer bias and optimizing extrinsic parameters. \emph{\scriptsize OAFE} refers to optimizing accelerometer bias and fixing extrinsic parameters.
\end{tablenotes}
\end{threeparttable}
\end{table}

Table \ref{table_average_error} indicates that the proposed approach outperforms all the other five approaches on 9 out of the 15 sequences. Among the rest six sequences, urban25, urban35 and urban37 do not contain turnings, as a result the proposed bidirectional trajectory computation does not come in handy on these sequences in our proposed approach. The accuracy of the proposed approach on the above three sequences is generally higher than FAOE, OAOE and the stereo VI-SLAM\cite{qin2019a}, and comparable with OAFE\cite{liu2019visual}, but slighterly lower than FAFE. That is because the manipulation described in Section \ref{ssec:bounded_marginalization} can cause information loss, given that when the trajectory does not contain turnings, the only difference between the proposed approach and FAFE lies in the utilization of the manipulation in Section \ref{ssec:bounded_marginalization}. However, in view of the good performance of the proposed approach on the other sequences with turnings where the bidirectional trajectory computation comes in handy, the benefit of the manipulation in Section \ref{ssec:bounded_marginalization} dramatically outweighs the cost. Generally speaking, the accuracy of the proposed approach is higher than \cite{liu2019visual}, as well as its adapted versions that deal with the accelerometer bias and extrinsic parameters differently.

\subsection{Absolute Trajectory Error (ATE) Comparison}

We also make an extensive comparison with more approaches, including the odometer-aided  VI-SLAM approaches \cite{Zhang2019vision}, \cite{zhang2019localization} and \cite{liu2019visual}, the stereo VI-SLAM system VINS-Fusion \cite{qin2019a}, the standard VIO \cite{li2013optimization}, and the batch and incremental stereo VI-SLAM approach ICE-BA\cite{liu2018ice}. The comparison is made in terms of absolute trajectory error (ATE), which is the rooted mean square error (RMSE) of the positions after a 6-DoF trajectory alignment with the ground truth. The experiments are conducted on the sequences urban26, urban28, urban38 and urban39, because only the ATEs of these four sequences are reported in the paper \cite{zhang2019localization}. The comparison results are shown in Table \ref{table_rmse}.

Table \ref{table_rmse} indicates that the proposed approach is more accurate than other approaches in terms of ATE on 3 out of the 4 sequences, and on the rest one sequence urban26, the accuracy of the proposed approach is second only to \cite{liu2019visual} by a small difference.

\subsection{Evaluation of Effects on Estimating Accelerometer Bias and Extrinsic Parameter}
\begin{figure}[htb]
  \centering
   \subfigure{
  \label{urban27_obscmp}
  \centering
  \includegraphics[width=7.9cm]{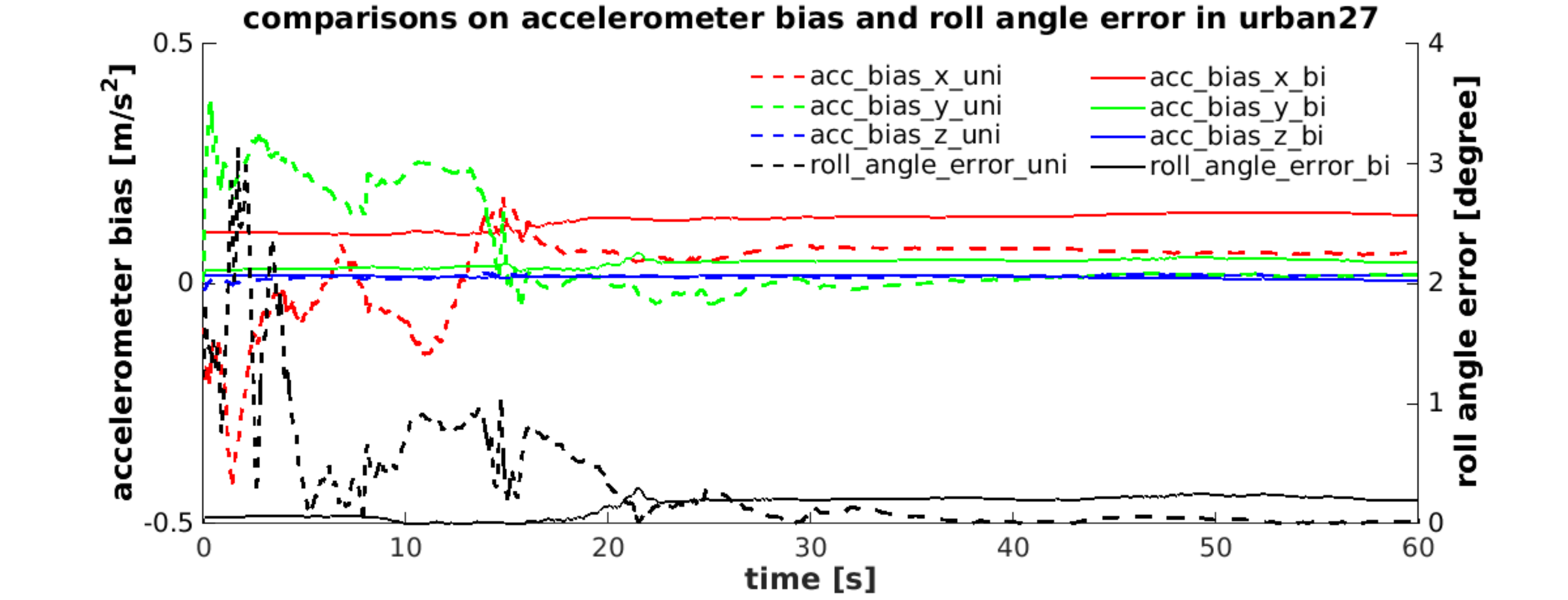}
  }
  \subfigure{
  \label{urban28_obscmp}
  \centering
  \includegraphics[width=7.9cm]{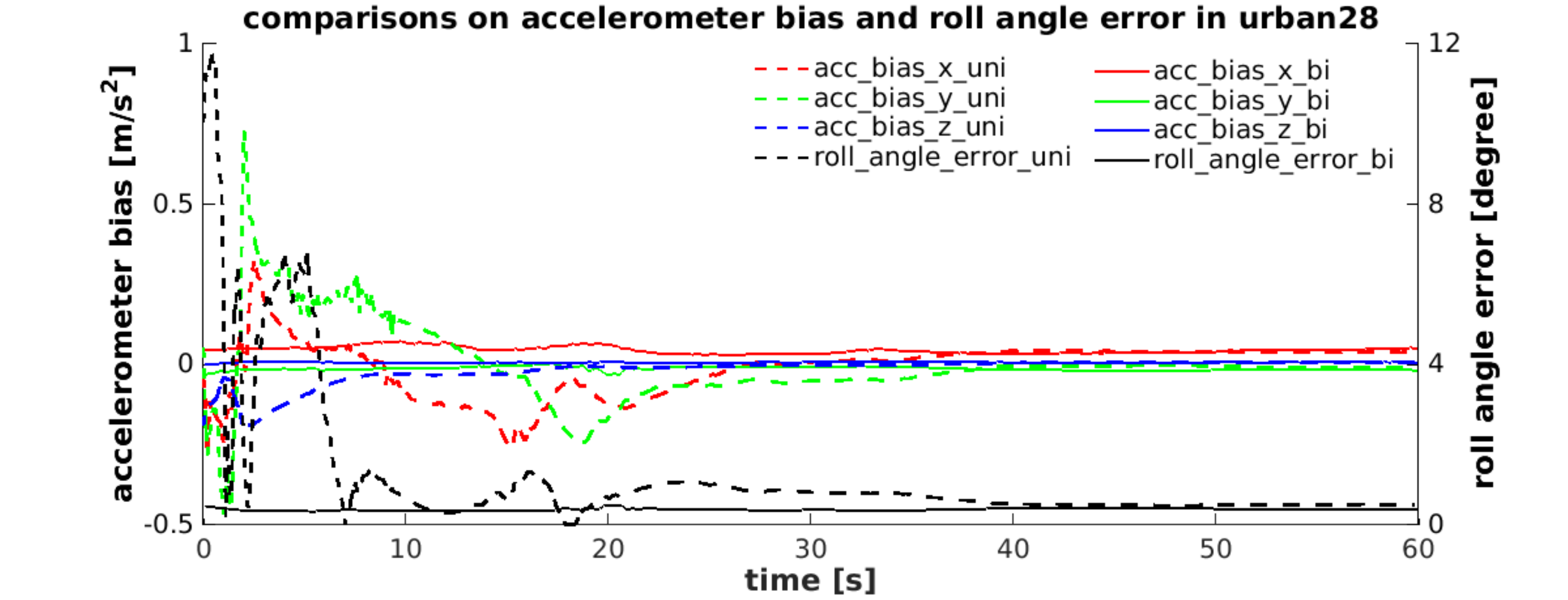}
  }
\caption{Comparison on estimated accelerometer bias and the estimation error in roll angle of $\mathbf{R}^b_c$. \emph{acc\_bias\_x\_uni}, \emph{acc\_bias\_y\_uni}, \emph{acc\_bias\_z\_uni} and \emph{roll\_angle\_error\_uni} are respectively the three components of accelerometer bias and the roll angle error estimated by the unidirectional computation method \emph{\footnotesize OAOE}.  \emph{acc\_bias\_x\_bi}, \emph{acc\_bias\_y\_bi}, \emph{acc\_bias\_z\_bi} and \emph{roll\_angle\_error\_bi} are the corresponding quantities estimated by our proposed bidirectional trajectory computation method. Here \emph{acc} refers to accelerometer, and \emph{uni} and \emph{bi} mean unidirectional and bidirectional trajectory computation respectively.}
\label{fig_comp}
\end{figure}

We examine the effects on estimating accelerometer bias and extrinsic parameter using the proposed bidirectional trajectory computation approach, in order to reveal why this approach improves the accuracy. The proposed approach is compared with the approach OAOE in Section \ref{ssec:average_positioning_error}, which optimizes accelerometer bias and extrinsic parameters from the beginning and only performs forward computation. Figure \ref{fig_comp} shows the comparison on estimated values of accelerometer bias and the estimation error in roll angle of $\mathbf{R}^b_c$ between the above two approaches, at the first turning in each sequence of urban27 and urban28. As same as in Section \ref{sec:observability}, here the system also starts at a dozen seconds before the first turning in each sequence instead of starting from the very beginning. Figure \ref{fig_comp} indicates that the estimation error in roll angle of $\mathbf{R}^b_c$ is much smaller using the proposed approach, and that the estimated value of accelerometer bias is more stable over time, which is more reasonable because the accelerometer bias is a slow time-varying quantity. Besides, both approaches can estimate the z-component of the accelerometer bias well. That is because both of the two unobservable directions of accelerometer bias before the first turning, which compose the orthogonal basis spanning over the 2D column space of $\mathbf{R}^{b_0}_w[\mathbf{g}^w]_{\times}$, have very small values in their respective z-components, in consideration of the fact that $\mathbf{R}^{b_0}_w$ is a rotation mainly around the Z axis for a ground vehicle and the gravity direction $\mathbf{g}^w$ is exactly along the Z axis. The ground-truth value of $\mathbf{R}^b_c$ is obtained offline, and the method to compute error in the roll angle is the same as that in Section \ref{sec:observability}.

\subsection{Computation of Real-time Trajectory}
\begin{figure}[htb]
  \centering
  \subfigure[Trajectory after 0 seconds]{
  \label{subfig_0min}
  \centering
  \includegraphics[width=4.0cm]{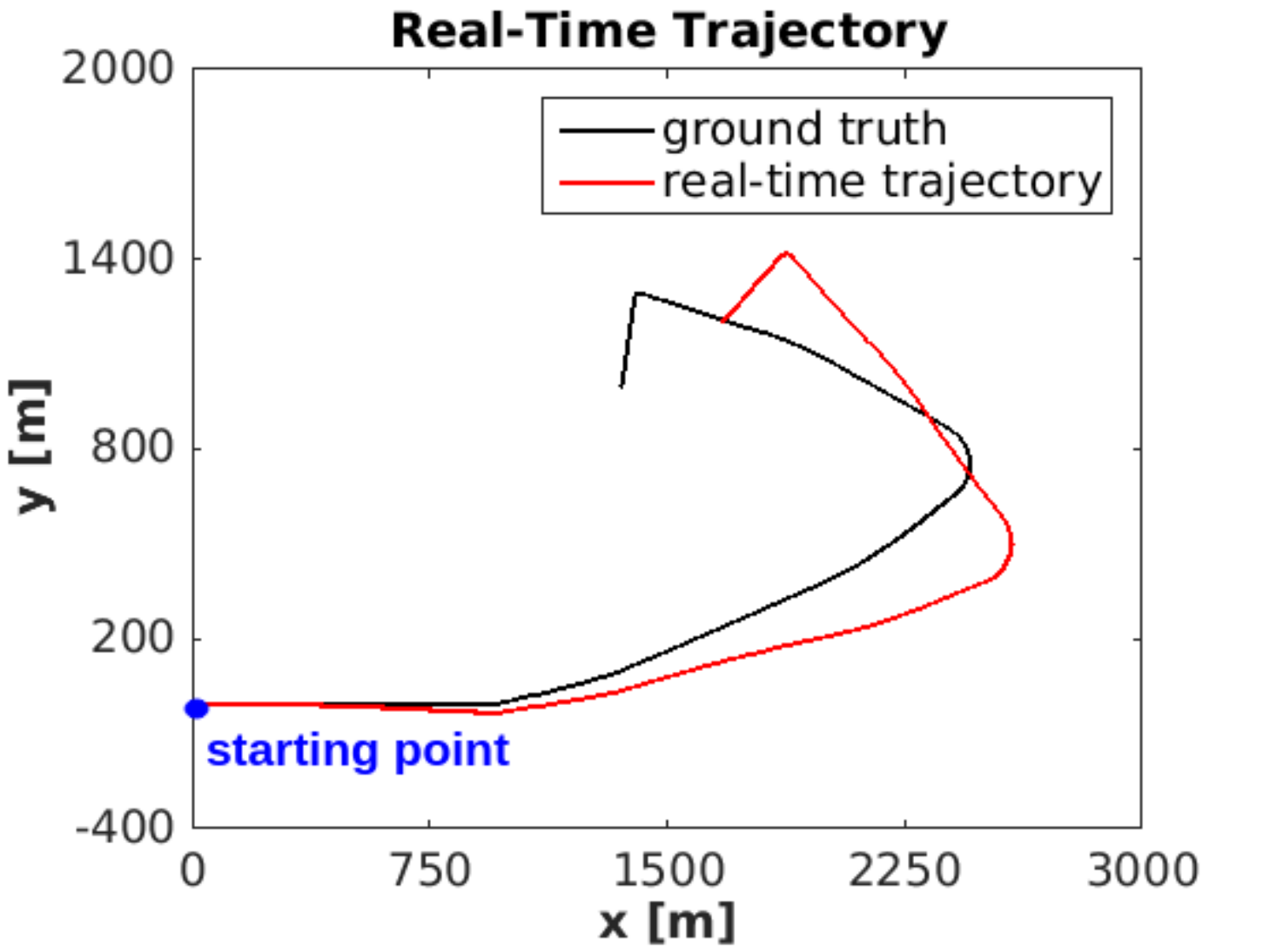}
  }
  \subfigure[Trajectory after 160 seconds]{
  \label{subfig_3min}
  \centering
  \includegraphics[width=4.0cm]{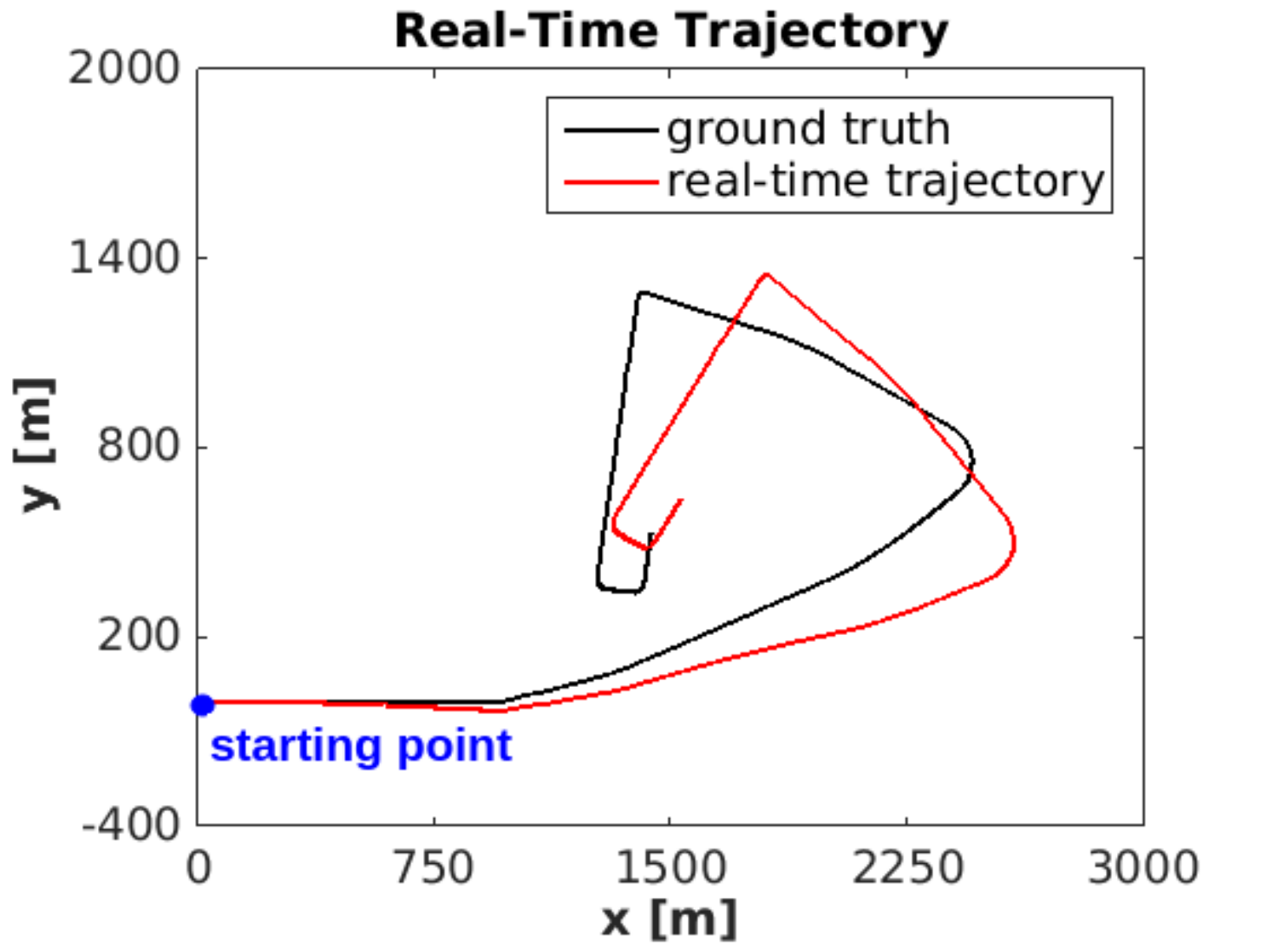}
  }
  \subfigure[Trajectory after 320 seconds]{
  \label{subfig_6min}
  \centering
  \includegraphics[width=4.0cm]{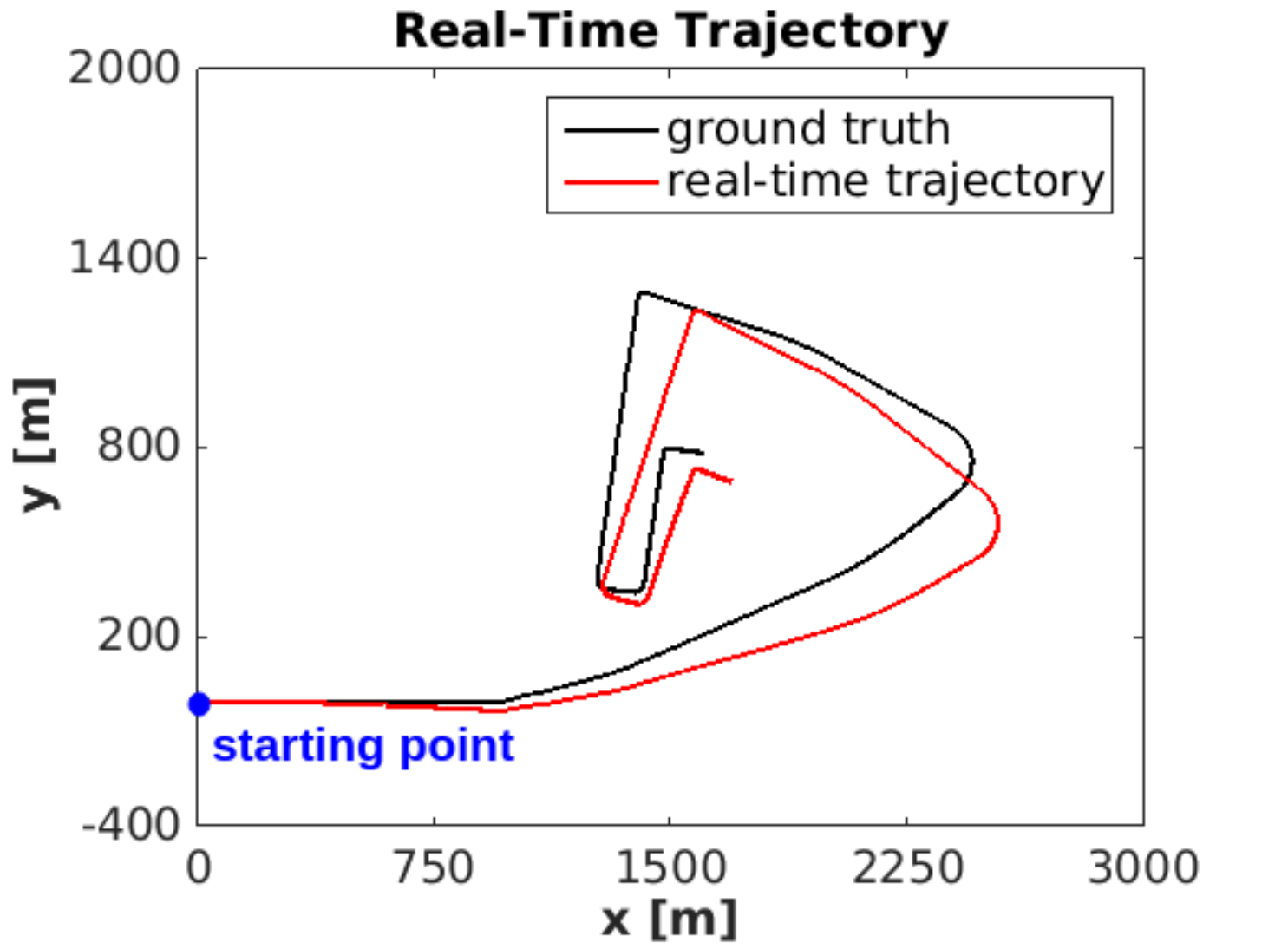}
  }
  \subfigure[Trajectory after 480 seconds]{
  \label{subfig_9min}
  \centering
  \includegraphics[width=4.0cm]{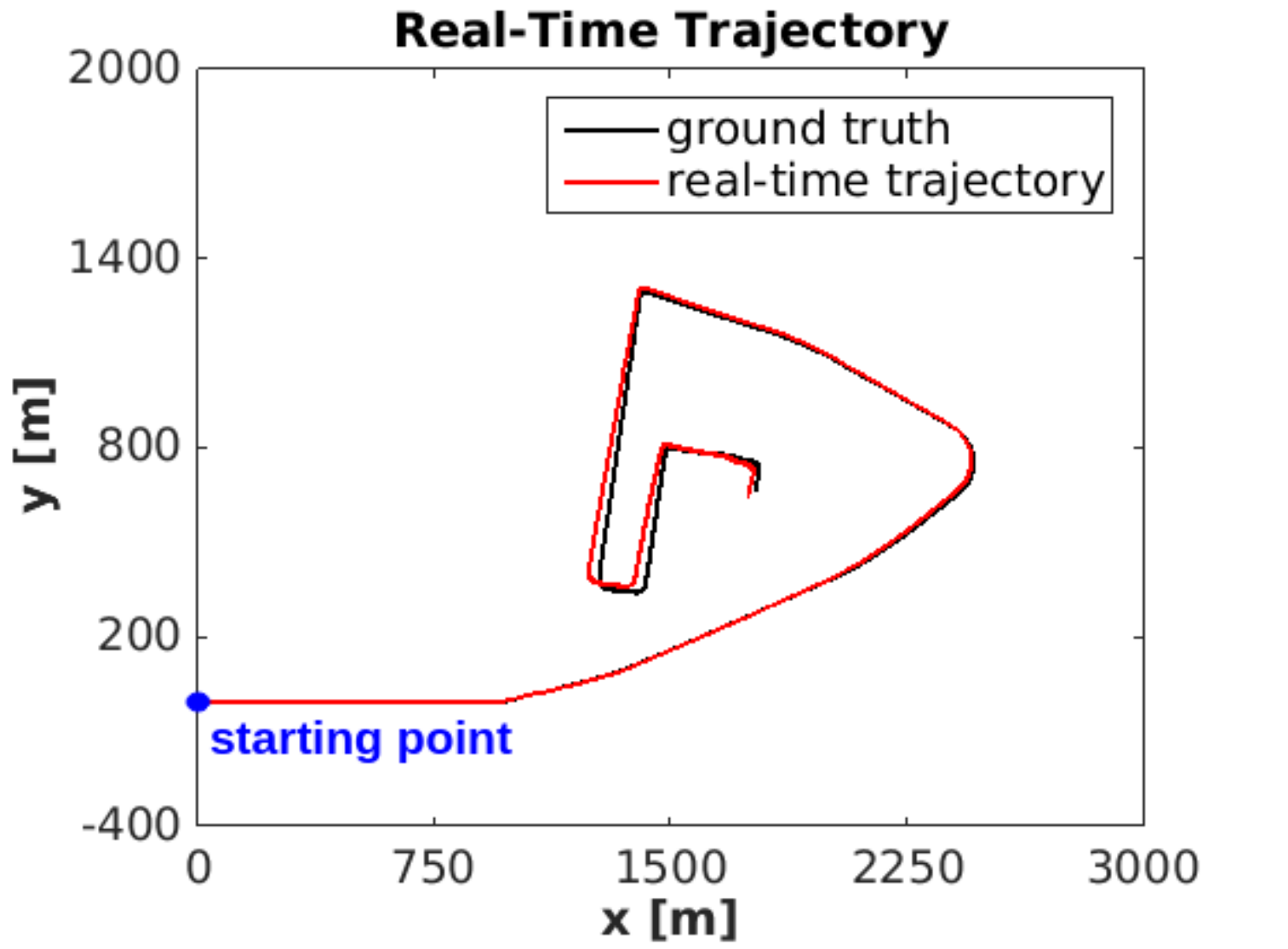}
  }
\caption{Real-time trajectories in urban33 at different moments.} 
\label{fig_realtime}
\end{figure}

To illustrate the effect of computing the real-time trajectory, we take the sequence urban33 for example. Figure \ref{subfig_0min}-\ref{subfig_9min} displays the real-time trajectories after 0, 160, 320 and 480 seconds since backward computation starts respectively, compared with the ground truth trajectory. We can see that as backward computation proceeds, the real-time trajectory becomes closer and closer to the ground truth trajectory gradually. 
\begin{table}[htb]
\scriptsize
\renewcommand{\arraystretch}{1.3}
\caption{Comparison of ATE (in meters) among different approaches}
\label{table_rmse}
\centering
\begin{threeparttable}
\begin{tabular}{*{8}{c}}
\bottomrule
Sequence&\multirow{3}{*}{proposed}&\multirow{3}{*}{\cite{liu2019visual}}&\multirow{3}{*}{\cite{zhang2019localization}}&\multirow{3}{*}{\cite{Zhang2019vision}}&VINS-&\multirow{3}{*}{\cite{li2013optimization}}&ICE-\\
(urban-)&&&&&Fusion&&BA\\
&&&&&\cite{qin2019a}&&\cite{liu2018ice}\\
\midrule
26 (4.0km)&12.0&\textbf{11.9}&14.8&16.1&22.5&32.8&22.1\\
28 (11.5km)&\textbf{15.4}&27.8&25.0&33.1&93.3&34.7&$\times$\\
38 (11.4km)&\textbf{11.8}&16.0&33.5&43.0&90.0&55.5&$\times$\\
39 (11.0km)&\textbf{7.5}&8.0&21.3&24.0&$\times$&33.4&$\times$\\
\bottomrule
\end{tabular}
\begin{tablenotes}
        \fontsize{8pt}{10pt}
		\item Here \emph{\scriptsize Proposed} means the proposed approach in this paper and '$\times$' means failure. \emph{\scriptsize Sequence} column firstly contains the sequence number which is \emph{\scriptsize urbanXX}, followed by the trajectory length in the parentheses. Results for \cite{zhang2019localization}, \cite{Zhang2019vision} and \cite{li2013optimization} are obtained from the reported results in \cite{zhang2019localization}. \emph{\scriptsize ATE} refers to absolute trajectory error.
\end{tablenotes}
\end{threeparttable}
\end{table}

\subsection{Comparison of Efficiency Against Batch and Incremental Solver}

We compare the efficiency of our proposed approach against the batch and incremental solver ICE-BA\cite{liu2018ice} by comparing the total consumption of time of computing the whole trajectory for each sequence, for which the results are shown in Table \ref{table_time}. From Table \ref{table_time} we can see that our proposed approach consumes far less time than ICE-BA\cite{liu2018ice}. From the above experiments we can draw that our proposed approach outperforms the incremental solver ICE-BA\cite{liu2018ice} in both accuracy and efficiency in KAIST Urban Data Set\cite{jeong2019complex}.
\begin{table}[htb]
\scriptsize
\renewcommand{\arraystretch}{1.3}
\caption{Comparison of time consumption (in seconds)}
\label{table_time}
\centering
\begin{threeparttable}
\begin{tabular}{*{6}{c}}
\bottomrule

\midrule
Sequence&urban25&urban26&urban27&urban28&urban29\\
\midrule
Proposed&107.4&\textbf{566.8}&\textbf{1133.5}&\textbf{2102.1}&\textbf{434.1}\\
ICE-BA\cite{liu2018ice}&\textbf{103.3}&967.9&3339.8&6940.5&1612.9\\
\midrule
Sequence&urban30&urban31&urban32&urban33&urban34\\
\midrule
Proposed&\textbf{1274.2}&\textbf{1113.6}&\textbf{1090.7}&\textbf{1335.9}&\textbf{659.2}\\
ICE-BA\cite{liu2018ice}&2093.3&3980.5&4789.7&6250.3&2352.1\\
\midrule
Sequence&urban35&urban36&urban37&urban38&urban39\\
\midrule
Proposed&\textbf{184.4}&\textbf{375.2}&\textbf{543.5}&\textbf{2159.8}&\textbf{1858.5}\\
ICE-BA\cite{liu2018ice}&572.4&905.8&1398.4&13349.6&7220.3\\
\bottomrule
\end{tabular}
\begin{tablenotes}
        \fontsize{8pt}{10pt}
		\item Here \emph{\scriptsize Proposed} means the proposed approach in this paper.
\end{tablenotes}
\end{threeparttable}
\end{table}

\section{CONCLUSION}
In this paper, we propose a bidirectional trajectory computation method for VI-SLAM aided with wheel encoder. Firstly, we perform an observability analysis on the degenerate case that an odometer-aided VI-SLAM system deployed on a car possibly encounters before the first turning. Secondly, we describe our proposed backward computation thread which refines the poses before the first turning, as well as the method to adjust the real-time trajectory. Experimental results show the higher accuracy of the whole trajectory, the correctly estimated parameters before the first turning, and the effects of real-time trajectory adjustment. Although in this paper wheel encoder is used, we also believe that the proposed bidirectional trajectory computation method can be applied on VI-SLAM systems that are not aided with wheel encoders as well.

\end{document}